\documentclass{article}
\usepackage{spconf,amsmath,graphicx,hyperref}
\usepackage{enumitem}

\makeatletter
\renewcommand\paragraph{\@startsection{paragraph}{4}{\z@}{1ex}{-1em}{\normalfont\normalsize\bfseries}}
\makeatother

\usepackage{cite}
\usepackage{algorithmic}
\usepackage{textcomp}

\usepackage{amsmath,graphicx}
\usepackage{amssymb}
\usepackage{booktabs} 
\usepackage{multirow}
\usepackage{xcolor}
\usepackage{tcolorbox}

\usepackage{pifont}
\makeatletter

\renewcommand\section{\@startsection{section}{1}{\z@}%
  {-2.2ex plus -.5ex minus -.2ex}{1.5ex plus .3ex}%
  {\normalfont\normalsize\bfseries\centering\MakeUppercase}}
\renewcommand\subsection{\@startsection{subsection}{2}{\z@}%
  {-2.0ex plus -.5ex minus -.2ex}{1.3ex plus .3ex}%
  {\normalfont\normalsize\bfseries}}
\makeatother
  
\definecolor{c1}{cmyk}{0,0.6175,0.8848,0.1490}
\definecolor{c2}{cmyk}{0.1127,0.6690,0,0.4431}
\definecolor{c3}{cmyk}{0.3081,0,0.7209,0.3255}
\definecolor{c4}{cmyk}{0.6765,0.2017,0,0.0667}
\definecolor{c5}{cmyk}{0,0.8765,0.7099,0.3647}
\definecolor{forestgreen}{HTML}{397727}

\definecolor{lightgreen}{RGB}{180,230,160} 
\definecolor{lightred}{RGB}{255,100,99} 

\usepackage{pgfplots}
\pgfplotsset{compat=1.17}
\usepackage{subcaption}
\usepackage{graphicx}

\let\OLDthebibliography\thebibliography
\renewcommand\thebibliography[1]{%
  \OLDthebibliography{#1}%
  \setlength{\parskip}{0pt}%
  \setlength{\itemsep}{0.5ex}%
}
\title{Deep TPC: Temporal-Prior Conditioning for Time Series Forecasting}

\name{Filippos Bellos$^{1}$, NaveenJohn Premkumar$^{1}$, Yannis Avrithis$^{2}$, Nam H. Nguyen$^{3}$, Jason J. Corso$^{1}$}
\address{
$^{1}$University of Michigan, Ann Arbor, MI, USA \\
$^{2}$Independent Scientist \\
$^{3}$Capital One, McLean, VA, USA \\
\{fbellos, naveenjp, jjcorso\}@umich.edu, yannis@avrithis.net, nam.nguyen@capitalone.com
}

\makeatletter
\def\ps@IEEEtitlepagestyle{%
  \def\@oddfoot{\hfill%
    \raisebox{-6pt}{
    \parbox[b]{\textwidth}{\footnotesize
    \copyright\ 2026 IEEE. Personal use of this material is permitted. Permission from IEEE must be obtained for all other uses, in any current or future media, including reprinting/republishing this material for advertising or promotional purposes, creating new collective works, for resale or redistribution to servers or lists, or reuse of any copyrighted component of this work in other works.\par
    }%
    }
  \hfill}%
  \def\@evenfoot{\@oddfoot}%
}
\makeatother

\begin{document}
%
\maketitle
\thispagestyle{IEEEtitlepagestyle}

\begin{abstract}
LLM–for–time series (TS) methods typically treat time shallowly, injecting positional or prompt-based cues once at the input of a largely frozen decoder, which limits temporal reasoning as this information degrades through the layers.
We introduce Temporal-Prior Conditioning (TPC), which elevates time to a first-class modality that conditions the model at multiple depths.
TPC attaches a small set of learnable time series tokens to the patch stream; at selected layers these tokens cross-attend to temporal embeddings derived from compact, human-readable temporal descriptors encoded by the same frozen LLM, then feed temporal context back via self-attention. 
This disentangles time series signal and temporal information while maintaining a low parameter budget. 
We show that by training only the cross-attention modules and explicitly disentangling time series signal and temporal information, TPC consistently outperforms both full fine-tuning and shallow conditioning strategies, 
achieving state-of-the-art performance in long-term forecasting across diverse datasets.
Code available at: \href{https://github.com/fil-mp/Deep_tpc}{github.com/fil-mp/Deep\_tpc}
\end{abstract}
\section{Introduction}
\label{sec:intro}

Large Language Models (LLMs) have revolutionized the field of natural language processing (NLP), demonstrating exceptional performance not only in traditional NLP tasks like text generation but also exhibiting significant potential in tasks demanding complex reasoning~\cite{wei2022chain,bellos-etal-2024-large}.

\begin{figure}[h]
  \centering
  \hspace*{-0.35cm} 
  \includegraphics[width=0.43\textwidth]{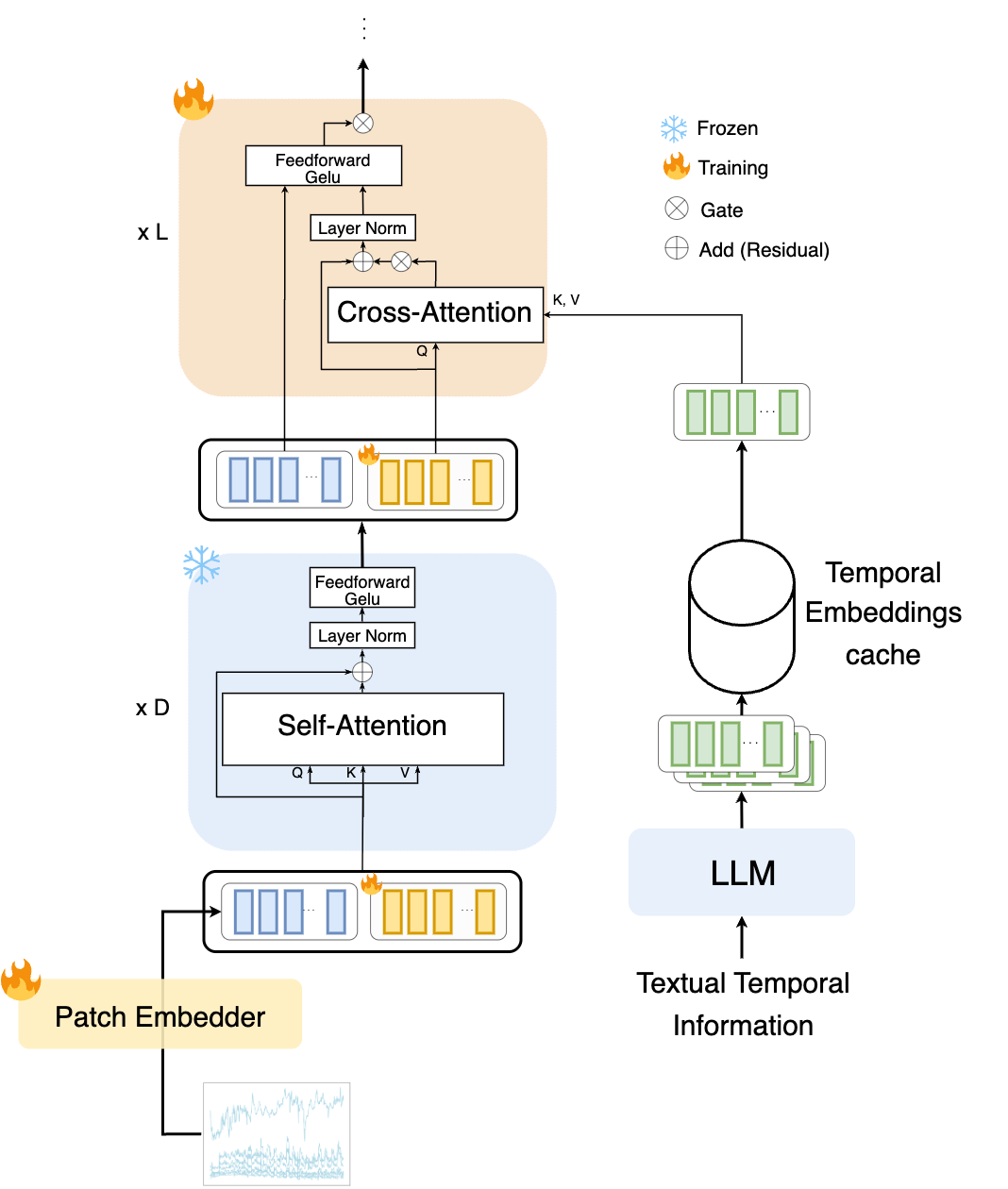}\vspace{-0.1mm} %
  \caption{
  TPC Overview. TPC modules are inserted at L layers of a frozen LLM. TS-tokens (yellow) mediate between time series patch embeddings (blue) and temporal embeddings (green) by performing cross-attention to gather temporal information, which is then distributed to patches via self-attention.
  }
  \label{pipeline}
  \vspace{-2mm}
\end{figure}

Beyond their original NLP domain, these models have achieved substantial progress in computer vision and other signal processing domains by enabling the creation of multimodal architectures capable of processing and synthesizing information across diverse modalities including text, images, and audio~\cite{li2023blip2}. 
Given this broad applicability and the unlocked multimodal reasoning capabilities, 
the community has begun exploring how LLMs might be leveraged for the time series domain~\cite{jin2023survey, 10889449}, a critical function underlying numerous real-world dynamic systems including energy load management~\cite{liu2023sadi}, climate modeling~\cite{schneider1974climate}, and traffic forecasting~\cite{DBLP:journals/corr/abs-2202-03630}.

Existing research on LLMs for time series tasks largely falls into two tracks, both predominantly shallow in how they condition the (mostly frozen) language decoder.
The first track utilizes LLMs to derive textual representations that inform time series modeling processes~\cite{wang2024a}. 
The second track, more relevant to time series forecasting, positions LLMs as central processing engines through either text-based or embedding-based representations.
Text-based approaches~\cite{ ansari2024chronos, wang2024b} convert numerical time series into textual tokens for direct LLM processing, though these methods face constraints related to sequence length and computational complexity. Alternatively, embedding-based approaches transform signals into continuous vector representations that are fed into pre-trained LLMs, frequently incorporating prompt optimization or data normalization techniques~\cite{zhou2023one, liu2024autotimes}.
Recent developments have focused on bridging time series embeddings with linguistic representations~\cite{pan2024sipllm},
and employing contrastive learning or reprogramming methodologies~\cite{sun2024test, jin2024timellm}. Additional works~\cite{10889449, Liu2023UniTimeAL}
introduce domain-specific vocabularies or utilize reconstruction-based pretraining approaches.

While both approaches operate on temporally ordered data, they generally treat temporal information implicitly and only at a shallow level. Text-based approaches rely on prefix prompting, statistical descriptors, or word embeddings~\cite{jin2024timellm}, where time is implicit in the sequence but not explicitly modeled. Embedding-based approaches usually attach positional encodings or normalization schemes; among them, AutoTimes~\cite{liu2024autotimes} extends this by injecting positional embeddings, yet this still restricts temporal information to the input stage. In all cases, time is handled as auxiliary metadata rather than as a distinct representational channel.

We argue that temporal information should be elevated to a first-class modality—complementing the signal modality itself and interacting with it throughout the reasoning stack.
Inspired by Multimodal Large Language Model (MLLM) architectures that rely on cross-attention and learnable tokens for robust cross-modal reasoning~\cite{li2023blip2,jaegle2022perceiverio,georgiou2025deepmlf}, we propose \textbf{Temporal-Prior Conditioning (TPC)}: a framework where specialized learnable time-series tokens (TS-tokens) disentangle the time series and temporal modalities and repeatedly interact with temporal embeddings across decoder layers. 

This modular design ensures that temporal priors persist across layers, rather than fading after a single injection step at the input level. 
Our experiments validate this hypothesis: TPC's deep temporal conditioning consistently outperforms full finetuning and shallow temporal integration strategies across most datasets.
\section{Method}

\subsection{Preliminaries}

\noindent\textbf{Problem formulation} We consider the standard long-term forecasting problem on multivariate time series.
Let $X \in \mathbb{R}^{N \times T}$ denote the observed history of $N$ variables over a lookback window of length $T$,
with the $i$-th series written as $X_i \in \mathbb{R}^{1 \times T}$.
Given this history, the goal is to predict the next $\tau$ time steps, denoted by
$Y \in \mathbb{R}^{N \times \tau}$.
For predictions $\hat{Y} \in \mathbb{R}^{N \times \tau}$ produced by a model $f(\cdot)$,
the forecasting objective is to minimize the mean squared error:
\begin{equation}
\min_{f} \;\; \frac{1}{N \cdot \tau} \sum_{i=1}^{N} \sum_{t=1}^{\tau}
\big( Y_{i,t} - \hat{Y}_{i,t} \big)^2.
\end{equation}

\noindent\textbf{Method overview} To solve this problem, we propose treating time as a first-class conditioning signal and injecting it deeply throughout the decoder.
Concretely, we introduce Temporal-Prior Conditioning (TPC) modules placed across multiple layers of a frozen decoder-only LLM with depth $D$.
We insert $L$ such modules at selected layers (empirically chosen for optimal accuracy–parameter trade-off), 
enabling temporal priors to interact with the backbone without modifying its parameters. Each input sequence to the decoder consists of (i) patch embeddings of the time series (learned from numeric windows and projected to the LM hidden size), concatenated with (ii) a small bank of learnable time-series tokens (TS-tokens) that travel with the patch stream.
Our design completely disentangles temporal and time series modalities: temporal embeddings and patch embeddings never directly interact.
Within each TPC module, only the TS-tokens perform cross-attention to temporal embeddings --- obtained by encoding compact, human-readable temporal prompts with the same frozen LLM to capture chronology, calendar effects, and seasonal regularities in the LLM's representation space. 
The TS-tokens then propagate this gathered temporal context to the patch stream through LLM's self-attention layers.

Practically, our approach keeps the LLM entirely frozen. We only train (a) the patch embedder, (b) the TS-tokens, (c) the parameters of the TPC modules (attention, gates, layer norm and feed-forward) and (d) the output linear projection. This preserves the computational efficiency of parameter-efficient methods while enabling layerwise interaction between time series and temporal priors, directly leveraging the LLM's pre-trained temporal reasoning capabilities without risking catastrophic forgetting.
\subsection{Input encoding}
\noindent\textbf{Time series encoding}
Following convention, for each input time series \(X_i \in \mathbb{R}^T\), we first apply reversible instance normalization (RevIN) to mitigate distribution shift: \(\tilde{X}_i = \text{RevIN}(X_i)\). We then divide \(\tilde{X}_i\) into \(P\) overlapping or non-overlapping patches of length \(L_p\): \(X_{P,i} \in \mathbb{R}^{P \times L_p}\),
where \(P = \left\lfloor\frac{T-L_p}{S}\right\rfloor + 2\), and \(S\) is the horizontal sliding stride.
To obtain the final embeddings, we apply a linear transformation to each patch: \(E_i = W_e X_{P,i} + b_e\), where \(E_i \in \mathbb{R}^{P \times d}\), \(W_e \in \mathbb{R}^{d \times L_p}\) is a learnable weight matrix, \(b_e \in \mathbb{R}^d\) is a learnable bias vector, and \(d\) is the embedding dimension of the target LLM.

\noindent\textbf{Time series tokens and input sequence} We introduce a bank of $n_f$ learnable TS-tokens
$\mathbf{X}_f^{(0)} \in \mathbb{R}^{n_f \times d}$ that persist through depth and mediate temporal conditioning.
The input to the frozen decoder is the concatenation (denoted by $\;\| \;$) of the patch embeddings and the fusion tokens:
$\mathbf{H}^{(0)} = \big[\, E_i \;\| \; \mathbf{X}_f^{(0)} \,\big] \in \mathbb{R}^{(P + n_f) \times d}.$
\subsection{Temporal embedding generation}

For each temporal span $p \in \{1,\dots,M\}$ (e.g. a lookback window or calendar slice),
we construct a compact textual description $x^{(p)}$ encoding its start and end timestamps along with the sampling granularity 
using a deterministic template 
(e.g. \textit{``This series spans 2017-01-01 00:00:00 to 2017-01-02 23:00:00"}).

These prompts are processed by the same frozen LLM used in the decoder pathway.
Let $\mathrm{Tok}(\cdot)$ denote the LLM tokenizer and $E_{\text{LLM}}$ its frozen input embedding matrix.
The tokenized prompt is mapped to embeddings as
$
\mathbf{X}^{(p)} = E_{\text{LLM}}\!\big(\mathrm{Tok}(x^{(p)})\big) \in \mathbb{R}^{L_p^{\text{txt}} \times d},
$
where $L_p^{\text{txt}}$ is the number of tokens in the textual prompt.

Passing these embeddings through the frozen LLM yields hidden states
$
\mathbf{H}^{(p)} = \mathrm{LLM}_{\text{frozen}}(\mathbf{X}^{(p)}) \in \mathbb{R}^{L_p^{\text{txt}} \times d},
$
and we take the final hidden state as the temporal embedding:
$
\mathbf{e}^{\text{temp}}_{p} = \mathbf{H}^{(p)}_{L_p^{\text{txt}}} \in \mathbb{R}^{d}.
$
Stacking across all spans produces a temporal embedding bank
$
\mathbf{E}_{\text{temp}} =
\big[\,\mathbf{e}^{\text{temp}}_{1}; \dots; \mathbf{e}^{\text{temp}}_{M}\,\big]
\in \mathbb{R}^{M \times d}.
$

Because these embeddings are generated by feeding textual prompts through the frozen LLM,
they reside in the same representational “language space” as the LLM’s own hidden states.
This ensures that temporal information is expressed in a form that the model can natively interpret and integrate,
rather than being injected as an arbitrary numeric encoding.
Since the LLM is frozen, $\mathbf{E}_{\text{temp}}$ can be precomputed once per span, cached, and reused during training and inference.

\subsection{Temporal conditioning}

The TPC framework conditions the patch embeddings on temporal priors via the TS-tokens. The pipeline begins with the frozen LLM's \emph{causal self-attention} layers, where patch embeddings and TS-tokens interact. Within each TPC module, \emph{gated cross-attention} then allows only the TS-tokens to query the temporal embeddings from $\mathbf{E}_{\text{temp}}$. The temporal information gathered by TS-tokens is subsequently propagated to patch embeddings through the next self-attention layers(fig.~\ref{pipeline}).

\noindent\textbf{Causal self-attention} Given input states $\mathbf{H}^{(l)}$ where
$
\mathbf{H}^{(l)} = \left[ E^{(l)} \;\|\; \mathbf{X}_{\text{temp}}^{(l)} \right]
\in \mathbb{R}^{(P+n_f)\times d}
$
at layer $l$, we compute query, key, and value projections:
$Q = \mathbf{H}^{(l)} W_Q$,
$K = \mathbf{H}^{(l)} W_K$,
$V = \mathbf{H}^{(l)} W_V$,
with $W_Q, W_K, W_V \in \mathbb{R}^{d \times d}$.
Causal masking $\mathcal{M}_{\text{causal}}$ ensures autoregressive flow along the patch sequence.
The update is
$
\tilde{\mathbf{H}}^{(l)}
= \mathrm{softmax} \left( QK^\top / \sqrt{d} + \mathcal{M}_{\text{causal}} \right) V,
$
followed by normalization and residual connections.
We write the result as
$
\tilde{\mathbf{H}}^{(l)} = \left[ \tilde{E}^{(l)} \;\|\; \tilde{\mathbf{X}}_{\text{temp}}^{(l)} \right].
$

\noindent\textbf{Gated cross-attention (TS-tokens $\to$ temporal embeddings)} Only the learnable TS-tokens $\tilde{\mathbf{X}}_{\text{temp}}^{(l)}$ attend to
the temporal embedding $\mathbf{E}_{\text{temp}} \in \mathbb{R}^{d}$.
This is enforced by a mask that blocks patch embeddings from accessing $\mathbf{E}_{\text{temp}}$.

Formally,
$Q_{\text{temp}} = \mathrm{Norm}(\tilde{\mathbf{X}}_{\text{temp}}^{(l)}) W_Q^t$,
$K_t = \mathbf{E}_{\text{temp}} W_K^t$,
$V_t = \mathbf{E}_{\text{temp}} W_V^t$,
with $W_Q^t, W_K^t, W_V^t \in \mathbb{R}^{d \times d}$.
The masked cross-attention output is
\begin{align}
\mathrm{CA}(\tilde{\mathbf{X}}_{\text{temp}}^{(l)}, \mathbf{E}_{\text{temp}}) =
\mathrm{softmax} \left( Q_{\text{temp}} K_t^\top / \sqrt{d} \right) V_t.
\end{align}

A learned gate $g_1^{(l)} = \sigma(a_1^{(l)})$, where $\sigma(\cdot)$ is the sigmoid function
and $a_1^{(l)}$ is a learned gating scalar, controls how much temporal information is injected:
$
\bar{\mathbf{X}}_{\text{temp}}^{(l)}
= \tilde{\mathbf{X}}_{\text{temp}}^{(l)}
+ g_1^{(l)} \cdot \mathrm{CA}(\tilde{\mathbf{X}}_{\text{temp}}^{(l)}, \mathbf{E}_{\text{temp}}).
$

\noindent\textbf{Gated feed-forward} The updated sequence is
$
\bar{\mathbf{H}}^{(l)}
= \big[\,\tilde{E}^{(l)} \;\|\; \bar{\mathbf{X}}_{\text{temp}}^{(l)}\,\big],
$
which is then passed through a gated feed-forward layer:
$
\mathbf{H}^{(l+1)}
= \bar{\mathbf{H}}^{(l)}
+ g_2^{(l)} \cdot \mathrm{FFN} \left( \mathrm{Norm}(\bar{\mathbf{H}}^{(l)}) \right),
$
with a second learned gate $g_2^{(l)} = \sigma(a_2^{(l)})$.
Gates are initialized to 0.5 (equal weighting).

\subsection{Next-token prediction}

\noindent\textbf{Forecasting objective} Following~\cite{liu2024autotimes}, our forecasting process adopts an autoregressive next-token prediction scheme, aligned with the pretraining objective of decoder-only LLMs. Given the normalized and patched time series inputs $E_i \in \mathbb{R}^{P \times d}$ together with the temporal priors injected by the TS-tokens $\mathbf{X}_{\text{temp}}$, the model autoregressively predicts the next $\tau$ time steps. After passing through $L$ TPC blocks, the final hidden states
$
\mathbf{H}^{(L)} =
\left[ E^{(L)} \;\|\; \mathbf{X}_{\text{temp}}^{(L)} \right]
\in \mathbb{R}^{(P+n_f) \times d}
$
encode fused representations of the signal and temporal context. Only the patch positions $E^{(L)} \in \mathbb{R}^{P \times d}$ are used for forecasting.

\noindent\textbf{Output projection} A learnable projection head maps each patch state to its next-step prediction. For the $p$-th patch token $E^{(L)}_{p} \in \mathbb{R}^{d}$, $\hat{x}_{p+1} = W_o E^{(L)}_{p} + b_o$, where $W_o \in \mathbb{R}^{L_p \times d}$, $b_o \in \mathbb{R}^{L_p}$ and $\hat{x}_{p+1} \in \mathbb{R}^{L_p}$ denotes the predicted values for the subsequent patch of length $L_p$.

\begin{table*}[!h]
\footnotesize
\centering
\begin{tabular}{lcccccccccccccc}
\toprule
\multirow{2}{*}{\sc{Method}} &
  \multicolumn{2}{c}{\sc{TPC (ours)}} &
  \multicolumn{2}{c}{\sc{AutoTimes}} &
  \multicolumn{2}{c}{\sc{OFA}} &
  \multicolumn{2}{c}{\sc{TimeLLM}} &
  \multicolumn{2}{c}{\sc{$S^2$IP-LLM}} &
  \multicolumn{2}{c}{\sc{PatchTST}} &
  \multicolumn{2}{c}{\sc{Dlinear}}    \\ \cmidrule{2-15}
&
  \sc{MSE} &
  \sc{MAE} &
  \sc{MSE} &
  \sc{MAE} &
  \sc{MSE} &
  \sc{MAE} &
  \sc{MSE} &
  \sc{MAE} &
  \sc{MSE} &
  \sc{MAE} &
  \sc{MSE} &
  \sc{MAE} &
  \sc{MSE} &
  \sc{MAE} \\ \midrule

{ETTh1} & \textbf{0.399} & \textbf{0.422} & \underline{0.409} & 0.438 & 0.415 & \underline{0.429} & 0.451 & 0.451 & 0.425 & 0.440 & 0.444 & 0.453 & 0.418 & 0.439 \\

{ETTh2} & \textbf{0.355} & \textbf{0.399} & 0.365 & 0.403 & 0.367 & \underline{0.402} & 0.355 & 0.398 & \underline{0.358} & 0.403 & 0.381 & 0.411 & 0.502 & 0.481 \\

{ETTm1} & \textbf{0.346} & \textbf{0.379} & 0.358 & \underline{0.382} & 0.355 & 0.386 & 0.349 & \underline{0.382} & \underline{0.347} & \underline{0.382} & 0.363 & 0.391 & 0.357 & 0.389 \\

{ETTm2} & \underline{0.265} & \textbf{0.323} & 0.281 & 0.330 & 0.265 & 0.328 & 0.271 & 0.332 & \textbf{0.261} & 0.326 & 0.267 & \underline{0.325} & 0.275 & 0.340 \\

{Weather} & 0.230 & \underline{0.267} & 0.243 & 0.278 & 0.237 & 0.270 & 0.230 & 0.268 & \underline{0.229} & \underline{0.267} & \textbf{0.225} & \textbf{0.264} & 0.248 & 0.300 \\

{Electricity} & \underline{0.166} & \underline{0.258} & 0.173 & 0.266 & 0.167 & 0.263 & 0.167 & 0.261 & 0.167 & 0.263 & \textbf{0.161} & \textbf{0.252} & \underline{0.166} & 0.263 \\

{Traffic} & \underline{0.394} & \underline{0.264} & 0.406 & 0.276 & 0.414 & 0.294 & 0.408 & 0.290 & 0.418 & 0.303 & \textbf{0.390} & \textbf{0.263} & 0.433 & 0.295 \\

{Solar} & \textbf{0.201} & \textbf{0.255} & 0.209 & \underline{0.258} & 0.263 & 0.335 & 0.263 & 0.335 & 0.243 & 0.270 & \underline{0.202} & 0.269 & 0.222 & 0.283 \\

\bottomrule

\end{tabular}
\caption{Long-term forecasting results for \{96, 192, 336, 720\} horizons. A lower value indicates a better performance.
All results are averaged from four forecasting horizons \{96, 192, 336, 720\}.  \underline{Underlined}: second best. \textbf{Bold}: best. We reproduced AutoTimes, TimeLLM, OFA, $S^2$IP-LLM, PatchTST, and Dlinear results using their official open-source implementations.}
\label{tab:few10}
\vspace{-1mm}

\end{table*}

\noindent\textbf{Training objective} During training, model parameters are optimized using mean squared error (MSE) between the predicted patch and the corresponding ground truth:
\begin{equation}
\mathcal{L}_{\text{MSE}} = \frac{1}{N \cdot L_p}
\sum_{i=1}^{N} \sum_{t=1}^{L_p}
\big( Y_{i,t} - \hat{Y}_{i,t} \big)^2,
\end{equation}
where $\mathbf{Y} \in \mathbb{R}^{N \times L_p}$ are the ground-truth values for the next patch and $\hat{\mathbf{Y}}$ the model’s predictions.
Because forecasting is autoregressive, it is sufficient to train the model for a single prediction length $L_p$: at inference time, multi-step horizons $\tau$ are reached by iteratively re-encoding and predicting one patch at a time.

\noindent\textbf{Autoregressive inference} At test time, the predicted patch $\hat{x}_{p+1}$ is appended to the observed history, re-segmented into patches and re-encoded by the patch encoder. This updated sequence is passed back into the model to generate the following patch. The process is repeated until the horizon of $\tau$ steps is reached.

\noindent\textbf{Parameter efficiency} The pretrained LLM backbone remains frozen. We only optimize
(i) the patch encoder $(W_e,b_e)$,
(ii) the TS-tokens $\mathbf{X}_{\text{temp}}$,
(iii) the TPC modules parameters and
(iv) the output head $(W_o,b_o)$.
Our method is efficient because it achieves better performance than full fine-tuning
while training only half or fewer of the parameters (fig.~\ref{tab:ablation}).
\section{Experiments}
\label{results}

In our experimental evaluation, we assess the effectiveness of the proposed TPC framework.  
We benchmark TPC against recent LLM-based forecasting methods as well as state-of-the-art Transformer and non-Transformer baselines on the long-term forecasting task. For all experiments, we adopt GPT-2 small as the frozen backbone LLM, ensuring efficiency, reproducibility and fair comparison with baselines.
Experimental configurations follow the unified pipeline of~\cite{wu2022timesnet}\footnote{https://github.com/thuml/Time-Series-Library}.

\noindent\textbf{Baselines} We compare against strong LLM-based approaches, including AutoTimes~\cite{liu2024autotimes}, which integrates temporal information as positional encodings, and $S^2$IP-LLM~\cite{pan2024sipllm}, which partially fine-tunes the backbone model. We further include state-of-the-art Transformer-based and non-Transformer methods, namely PatchTST~\cite{nie2022time} and DLinear~\cite{Zeng2022AreTE}.

\noindent\textbf{Results} Table~\ref{tab:few10} shows that TPC outperforms LLM-based methods in almost all cases for both MSE and MAE, validating the benefit of disentangling temporal priors rather than relying on positional embeddings or prompt cues. Compared to Transformer-based PatchTST, TPC achieves state-of-the-art results on several datasets, while remaining competitive on the rest, being second-best in most cases.

\noindent\textbf{Ablation study} To better demonstrate the contribution of our design choices, we conduct ablations on the role of treating time as an explicit modality but also isolating the effect of deep temporal conditioning from parameter count. Specifically, we compare our method against shallow temporal integration (positional embeddings) trained with various strategies, including full and partial fine-tuning that match or exceed TPC's parameter budget. Specifically, we compare our method against several alternatives:
\begin{itemize}[noitemsep,topsep=0pt,partopsep=0pt, parsep=0pt, leftmargin=*]
    \item \emph{Positional Embeddings.} Injecting temporal information through additive positional embeddings, following the AutoTimes~\cite{liu2024autotimes} approach.
    \item \emph{Prefix-prompts} Concatenating temporal embeddings directly with the patch embeddings at the input layer, without maintaining a separate temporal channel.
    \item \emph{Full Fine-Tuning.} Fine-tuning all layers of the backbone LLM model.
    \item \emph{Partial Fine-Tuning.} Fine-tuning the same number of self-attention layers as the number of trained layers in TPC (TPC modules), in order to match parameter count.
    \item \emph{LoRA Fine-Tuning.} Adapting the LLM with low-rank adaptation technique (LoRA).
\end{itemize}
As shown in Table~\ref{tab:ablation}, AutoTimes (positional embeddings) serves as the most effective prior method for temporal modeling (better than prefix prompts). All fine-tuning variants are therefore applied on AutoTimes.
{
\setlength{\textfloatsep}{6pt} 
\setlength{\intextsep}{6pt}
\begin{table}[!h]
\footnotesize
\vspace{-3mm}

\centering
\resizebox{\columnwidth}{!}{
\begin{tabular}{lcccccc}
\toprule
\multirow{2}{*}{\sc{Method}} & \multicolumn{2}{c}{ETTh1} & \multicolumn{2}{c}{ETTm1} & \multicolumn{2}{c}{Weather} \\ \cmidrule{2-7}
 & \sc{MSE} & \sc{MAE} & \sc{MSE} & \sc{MAE} & \sc{MSE} & \sc{MAE} \\ \midrule
\textbf{TPC (ours)}            & \textbf{0.399} & \textbf{0.422} & \textbf{0.346} & \textbf{0.379} & \textbf{0.230} & \textbf{0.267} \\
Pos. Embeddings             & 0.409 & 0.438 & 0.358 & 0.382 & 0.243 & 0.278 \\
Prefix-prompts                  & 0.414 & 0.441 & \underline{0.352} & 0.390 & 0.243 & 0.279 \\
Full Fine-Tuning               & \underline{0.404} & 0.435 & 0.357 & \underline{0.381} & 0.235 & 0.273 \\
Partial Fine-Tuning            & 0.407 & 0.437 & 0.365 & 0.393 & 0.234 & \underline{0.270} \\
LoRA                           & 0.408 & \underline{0.432} & 0.358 & 0.382 & \underline{0.232} & \textbf{0.267} \\
\bottomrule
\end{tabular}}
\caption{Ablation study on temporal modality and training strategies across diverse datasets. \underline{Underlined}: second best. \textbf{Bold}: best}
\label{tab:ablation}
\vspace{-3mm}

\end{table}
}

This comparison isolates the effect of treating time as a disentagled first-class modality. Our TPC achieves the best balance by surpassing both full and partial fine-tuning with the same or fewer trainable parameters.

\section{Discussion}
TPC demonstrates that there is significant potential in treating time as a distinct modality and moving beyond shallow temporal injection, with deep conditioning achieving superior results across diverse datasets for long-term time series forecasting. We see two natural directions for extending this work. First, we plan to augment the input representation by incorporating word embeddings alongside the patch embeddings, enabling richer alignment between symbolic and numeric information. Second, we will investigate the use of alternative LLMs (e.g., LLaMA) to assess model-agnostic generalization and potential performance gains.

\section{Acknowledgments}
This research was funded, in part, by the U.S. Government, under Agreement No. 1AY2AX000062 and AFOSR under No. FA2386-25-1-4064. The views and conclusions contained in this document are those of the authors and should not be interpreted as representing the official policies, either expressed or implied, of the U.S. Government.

\bibliographystyle{IEEEbib}
\bibliography{strings,refs}

@InProceedings{C2,
  author = 	 "Jones, C.D. and Smith, A.B. and Roberts, E.F.",
  title =        "Article Title",
  booktitle =        "Proceedings Title",
  organization = "IEEE",
  year = 	 "2003",
  volume = 	 "II",
  pages = 	 "803-806"
}

@inproceedings{jin2024timellm,
title={Time-{LLM}: Time Series Forecasting by Reprogramming Large Language Models},
author={Ming Jin and Shiyu Wang and Lintao Ma and others},
booktitle={The Twelfth International Conference on Learning Representations},
year={2024},
url={https://openreview.net/forum?id=Unb5CVPtae}
}

@inproceedings{
zhou2023one,
title={One Fits All: Power General Time Series Analysis by Pretrained {LM}},
author={Tian Zhou and Peisong Niu and Xue Wang and Liang Sun and Rong Jin},
booktitle={Thirty-seventh Conference on Neural Information Processing Systems},
year={2023},
url={https://openreview.net/forum?id=gMS6FVZvmF}
}

@inproceedings{nie2022time,
title={A Time Series is Worth 64 Words:  Long-term Forecasting with Transformers},
author={Yuqi Nie and Nam H Nguyen and Phanwadee Sinthong and Jayant Kalagnanam},
booktitle={International Conference on Learning Representations},
year={2023}
}

@inproceedings{
pan2024sipllm,
title={\$S{\textasciicircum}2\${IP}-{LLM}: Semantic Space Informed Prompt Learning with {LLM} for Time Series Forecasting},
author={Zijie Pan and others},
booktitle={Forty-first International Conference on Machine Learning},
year={2024},
url={https://openreview.net/forum?id=qwQVV5R8Y7}
}

@inproceedings{wu2022timesnet,
  title={Timesnet: Temporal 2d-variation modeling for general time series analysis},
  author={Wu, Haixu and others},
  booktitle={International Conference on Learning Representations},
  year={2023}
}

@inproceedings{bellos-etal-2024-large,
    title = "Can Large Language Models Reason About Goal-Oriented Tasks?",
    author = "Bellos, Filippos and others",
    booktitle = "Proceedings of the First edition of the Workshop on the Scaling Behavior of Large Language Models (SCALE-LLM 2024)",
    month = mar,
    year = "2024",
    address = "St. Julian{'}s, Malta",
    publisher = "Association for Computational Linguistics",
    url = "https://aclanthology.org/2024.scalellm-1.3",
    pages = "24--34",
}

@inproceedings{Zeng2022AreTE,
  title={Are Transformers Effective for Time Series Forecasting?},
  author={Ailing Zeng and Muxi Chen and Lei Zhang and Qiang Xu},
  booktitle={Proceedings of the AAAI Conference on Artificial Intelligence},
  year={2023}
}

@inproceedings{
sun2024test,
title={{TEST}: Text Prototype Aligned Embedding to Activate {LLM}'s Ability for Time Series},
author={Chenxi Sun and others},
booktitle={The Twelfth International Conference on Learning Representations},
year={2024},
url={https://openreview.net/forum?id=Tuh4nZVb0g}
}

@INPROCEEDINGS{10889449,
  author={Bellos, Filippos and Nguyen, Nam H. and Corso, Jason J.},
  booktitle={ICASSP 2025 - 2025 IEEE International Conference on Acoustics, Speech and Signal Processing (ICASSP)}, 
  title={VITRO: Vocabulary Inversion for Time-series Representation Optimization}, 
  year={2025},
  volume={},
  number={},
  pages={1-5},
  doi={10.1109/ICASSP49660.2025.10889449}}

@article{jin2023survey,
  title={A Survey on Graph Neural Networks for Time Series: Forecasting, Classification, Imputation, and Anomaly Detection},
  author={Ming Jin and Huan Yee Koh and Qingsong Wen and others},
  journal={IEEE transactions on pattern analysis and machine intelligence},
  year={2023},
  volume={PP},
  url={https://api.semanticscholar.org/CorpusID:259501265}
}

@inproceedings{liu2023sadi,
  title={SADI: A Self-Adaptive Decomposed Interpretable Framework for Electric Load Forecasting Under Extreme Events},
  author={Liu, Hengbo and others},
  booktitle={IEEE International Conference on Acoustics, Speech and Signal Processing},
  year={2023},
}

@article{schneider1974climate,
  title={Climate modeling},
  author={Schneider, Stephen H and Dickinson, Robert E},
  journal={Reviews of Geophysics},
  volume={12},
  number={3},
  pages={447--493},
  year={1974},
  publisher={Wiley Online Library}
}

@article{DBLP:journals/corr/abs-2202-03630,
  author       = {Yihong Tang and
                  Ao Qu and
                  Andy H. F. Chow and
                  others},
  title        = {Domain Adversarial Spatial-Temporal Network: {A} Transferable Framework
                  for Short-term Traffic Forecasting across Cities},
  journal      = {CoRR},
  volume       = {abs/2202.03630},
  year         = {2022},
  url          = {https://arxiv.org/abs/2202.03630},
  eprinttype    = {arXiv},
  eprint       = {2202.03630},
  bibsource    = {dblp computer science bibliography, https://dblp.org}
}

@inproceedings{wei2022chain,
author = {Wei, Jason and others},
title = {Chain-of-thought prompting elicits reasoning in large language models},
year = {2024},
isbn = {9781713871088},
publisher = {Curran Associates Inc.},
address = {Red Hook, NY, USA},
booktitle = {Proceedings of the 36th International Conference on Neural Information Processing Systems},
articleno = {1800},
numpages = {14},
location = {New Orleans, LA, USA},
series = {NIPS '22}
}

@article{Liu2023UniTimeAL,
  title={UniTime: A Language-Empowered Unified Model for Cross-Domain Time Series Forecasting},
  author={Xu Liu and others},
  journal={Proceedings of the ACM Web Conference 2024},
  year={2023},
  url={https://api.semanticscholar.org/CorpusID:264146377}
}

@inproceedings{li2023blip2,
  title={BLIP-2: Bootstrapping Language-Image Pre-training with Frozen Image Encoders and Large Language Models},
  author={Li, Junnan and Li, Dongxu and others},
  booktitle={Proceedings of the 40th International Conference on Machine Learning (ICML)},
  pages={19730--19742},
  year={2023},
  organization={PMLR}
}

@inproceedings{jaegle2022perceiverio,
  title     = {Perceiver IO: A General Architecture for Structured Inputs \& Outputs},
  author    = {Jaegle, Andrew and others},
  booktitle = {International Conference on Learning Representations (ICLR)},
  year      = {2022},
  note      = {Spotlight},
  url       = {https://openreview.net/forum?id=fILj7WpI-g}
}

@article{georgiou2025deepmlf,
  title   = {DeepMLF: Multimodal language model with learnable tokens for deep fusion in sentiment analysis},
  author  = {Georgiou, Efthymios and Katsouros, Vassilis and others},
  journal = {arXiv preprint arXiv:2504.11082},
  year    = {2025}
}

@inproceedings{liu2024autotimes,
  author = {Liu, Yong and Qin, Guo and Huang, Xiangdong and others},
  title = {AutoTimes: Autoregressive Time Series Forecasters via Large Language Models},
  booktitle = {Proceedings of the 38th International Conference on Neural Information Processing Systems},
  year = {2024},
  url = {https://proceedings.neurips.cc/paper_files/paper/2024/hash/dcf88cbc8d01ce7309b83d0ebaeb9d29-Abstract-Conference.html},
  publisher = {NeurIPS},
  pages = {122154--122184}
}

@inproceedings{wang2024b,
  title={From news to forecast: Integrating event analysis in llm-based time series forecasting with reflection},
  author={Wang, Xinlei and Feng, Maike and Qiu, Jing and Gu, Jinjin and others},
  booktitle={Advances in Neural Information Processing Systems},
  volume={37},
  pages={58118--58153},
  year={2024}
}

@inproceedings{wang2024a,
  title={CTPD: Cross-Modal Temporal Pattern Discovery for Enhanced Multimodal Electronic Health Records Analysis},
  author={Wang, Fuying and Wu, Feng and others},
  booktitle={Findings of the Association for Computational Linguistics: ACL 2025},
  pages={6783--6799},
  year={2025}
}

@article{ansari2024chronos,
  title={Chronos: Learning the Language of Time Series},
  author={Ansari, Abdul Fatir and Stella, Lorenzo and Turkmen, Caner and others},
  journal={Transactions on Machine Learning Research},
  year={2024},
}

\end{document}